\pdfoutput=1

\documentclass{article}

\usepackage[final]{neurips_2020}

\usepackage[utf8]{inputenc} 
\usepackage[T1]{fontenc}    
\usepackage{hyperref}       
\usepackage{url}            
\usepackage{booktabs}       
\usepackage{amsfonts}       
\usepackage{nicefrac}       
\usepackage{microtype}      

\usepackage{graphicx}
\usepackage{subfigure}
\usepackage{amsmath,amssymb,dsfont,mathrsfs}
\usepackage{tikz}
\usepackage{algorithm}
\usepackage{algorithmic}
\usetikzlibrary{calc}
\usetikzlibrary{arrows.meta}
\usepackage{enumitem}

\usepackage{xr-hyper}
\usepackage{fancyvrb}
\DefineVerbatimEnvironment{rawsrc}{BVerbatim}{fontsize=\small,baselinestretch=1.0}

\newcounter{nbdrafts}
\makeatletter
\newcommand{\checknbdrafts}{
\ifnum \thenbdrafts > 0
\@latex@warning@no@line{*WARNING* The document contains \thenbdrafts \space draft note(s)}
\fi}

\makeatother

\def\argmin{\operatornamewithlimits{argmin}}

\def\StateSpace{{\cal S}}
\def\ActionSpace{{\cal A}}
\def\GoalSpace{{\cal G}}
\newcommand{\LL}{\mathcal{L}}

\title{Multi-task Reinforcement Learning \\ with a Planning Quasi-Metric}

\author{
  Vincent Micheli\thanks{Work done while at EPFL.} \\
  University of Geneva \\
  \texttt{micheli-vincent@outlook.fr} \\
  \And
  Karthigan Sinnathamby\footnotemark[1] \\
  Visium \\
  \texttt{karthigan.sinnathamby@visium.ch} \\
  \AND
  François Fleuret\thanks{Work done while at Idiap/EPFL.} \\
  University of Geneva \\
  \texttt{francois.fleuret@unige.ch} \\
}

\begin{document}

\maketitle

\begin{abstract}
We introduce a new reinforcement learning approach combining a
planning quasi-metric (PQM) that estimates the number of steps
required to go from any state to another, with task-specific
``aimers'' that compute a target state to reach a given goal. This
decomposition allows the sharing across tasks of a task-agnostic model
of the quasi-metric that captures the environment's dynamics and can
be learned in a dense and unsupervised manner. We achieve
multiple-fold training speed-up compared to recently published methods
on the standard bit-flip problem and in the MuJoCo robotic arm
simulator.
\end{abstract}

\section{Introduction}

We are interested in devising a new approach to reinforcement learning
to solve multiple tasks in a single environment, and to learn
separately the dynamic of the environment and the definitions of goals
in it. A simple example would be a 2d maze, where there could be two
different sets of tasks: reach horizontal coordinate $x$ or reach
vertical coordinate $y$. Learning the spatial configuration of the
maze would be useful for both sets of tasks.

Our approach relies on task-specific models which, given a starting
state $s$, and a goal $g$ which is a {\it set} of states, compute a
``target state'' $s' \in g$. We dubbed these models {\bf aimers} (see
\S~\ref{sec:aimer}) and we stress that they are not designed to
compute the series of actions to go from $s$ to $s'$, but only $s'$
itself. Given this target state, the planning {\it per se} is computed
using a model of a {\bf quasi-metric} between states (see
\S~\ref{sec:plann-quasi-metr}). This latter model is task agnostic and
can be re-used from one to another. This decomposition allows to
transfer the modeling of the world dynamic captured by the quasi
metric, and to limit the task-specific learning to the aimers, which
are lighter models trainable with very few observations as
demonstrated in the experimental section (\S~\ref{sec:experiments}).

The idea of a quasi-metric between states is the natural extension of
recent works, starting with the Universal Value Function
Approximators~\citep{schaul15} which introduced the notion that
learning the reward function can be done without a single privileged
goal, and then extended with the Hindsight Experience
Replay~\citep{arxiv-1707.01495} that introduced the idea that goals do
not have to be pre-defined but can be picked arbitrarily. Combined
with a constant negative reward, this leads naturally to a metric
where states and goal get a more symmetric role, which departs from
the historical and classical idea of accumulated reward.

The long-term motivation of our approach is to
segment the policy of an agent into a life-long learned quasi-metric,
and a collection of task-specific easy-to-learn aimers. These
aimers would be related to high-level imperatives for a biological
system, triggered by low-level physiological necessities (``eat'',
``get warmer'', ``reproduce'', ``sleep''), and high-level operations
for a robot (``recharge battery'', ``pick up boxes'', ``patrol'',
etc.). Additionally, the central role of a metric where the heavy lifting
takes place provides a powerful framework to develop hierarchical
planning, curiosity strategies, estimators of performance, etc.


\begin{figure}[ht!]
\begin{minipage}{0.4\textwidth}
\center
\vspace*{12pt}
\makebox[\textwidth][c]{%
\begin{tikzpicture}
\draw[draw=none] (1.6, -1.4) rectangle (7.2, 3.);

\node[inner sep=1pt] (wall) at (3., 0.5) {
  \begin{tikzpicture}
    \draw[use as bounding box,draw=none,fill=black] (1, -1) rectangle ++(0.2, 2.9);
    \draw[draw=none,fill=black] (0, 1.7) rectangle ++(1.1, 0.2);
  \end{tikzpicture}
};

\node (G) at (5.1, 0.9) {
  \begin{tikzpicture}
    \draw[fill=black!30,draw=none] plot[smooth cycle,tension=1] coordinates {(1.9, 1) (3.3, 1.5) (2.2,  2)};
  \end{tikzpicture}
};

\node at ($(G)+(-.1, -.05)$) {$g$};

\coordinate (s1) at (3.8, 2.6);
\draw[draw=none,fill=red] (s1) circle (0.05) node[above left] {$s_1$};
\coordinate (s1p) at ($(G)+(-0.36, 0.45)$);
\draw[draw=none,fill=red] (s1p) circle (0.05) node[above right,xshift=0pt,yshift=3pt] {$s_1' = h^{\GoalSpace}(s_1, g)$};
\draw[dashed,very thick,red,shorten >=2.0pt,shorten <=2.0pt] (s1.center) -- (s1p);
\coordinate (s2) at (2.25, 0.25);
\draw[draw=none,fill=blue] (s2) circle (0.05) node[above left] {$s_2$};
\coordinate (s2p) at ($(G)+(-0.67, -0.44)$);
\draw[draw=none,fill=blue] (s2p) circle (0.05) node[below right,xshift=-4pt,yshift=-3pt] {$s_2' = h^{\GoalSpace}(s_2, g)$};
\draw[dashed,very thick,blue,shorten >=2.0pt,shorten <=2.0pt] (s2.center) -- (wall.south west) -- (wall.south east) -- (s2p);

\end{tikzpicture}
}

\caption{Given a starting state $s$ and a goal $g \in \GoalSpace$, the
  aimer $h^{\GoalSpace}$ computes a target state $s'$ which is the closest
  state in $g$, that is $s' = h^{\GoalSpace}(s, g) \in g$ that
  minimizes the length $\min_a f(s, s'\!, a)$ of the dashed path to go
  from $s$ to $s'$ under the environment dynamics.}\label{fig:aimer}
\end{minipage}
\hspace*{\stretch{1}}
\begin{minipage}{0.55\textwidth}

\makebox[\textwidth][c]{%
\includegraphics[height=2.25cm,trim=320 50 550 120,clip]{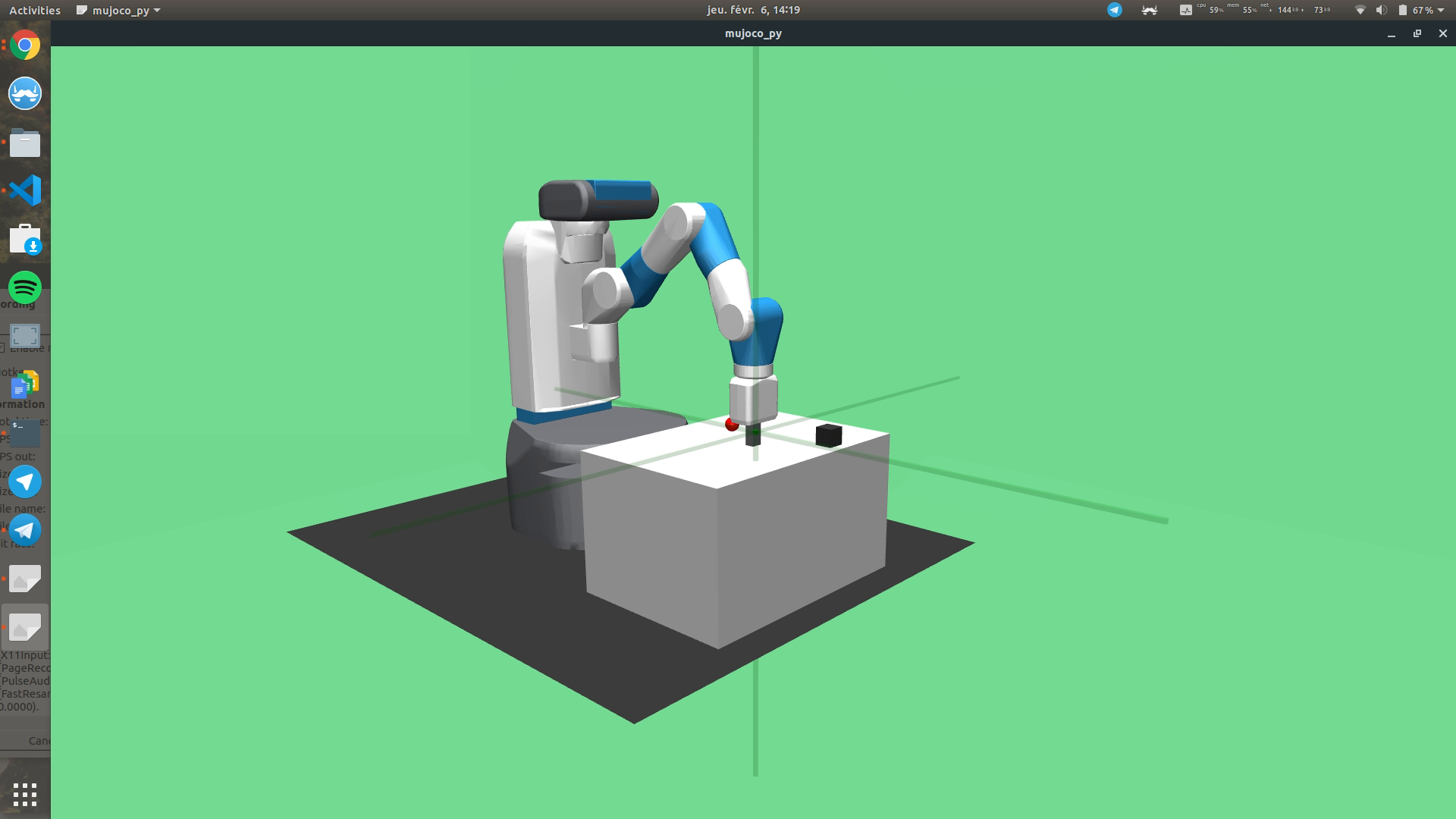}
\includegraphics[height=2.25cm,trim=320 50 550 120,clip]{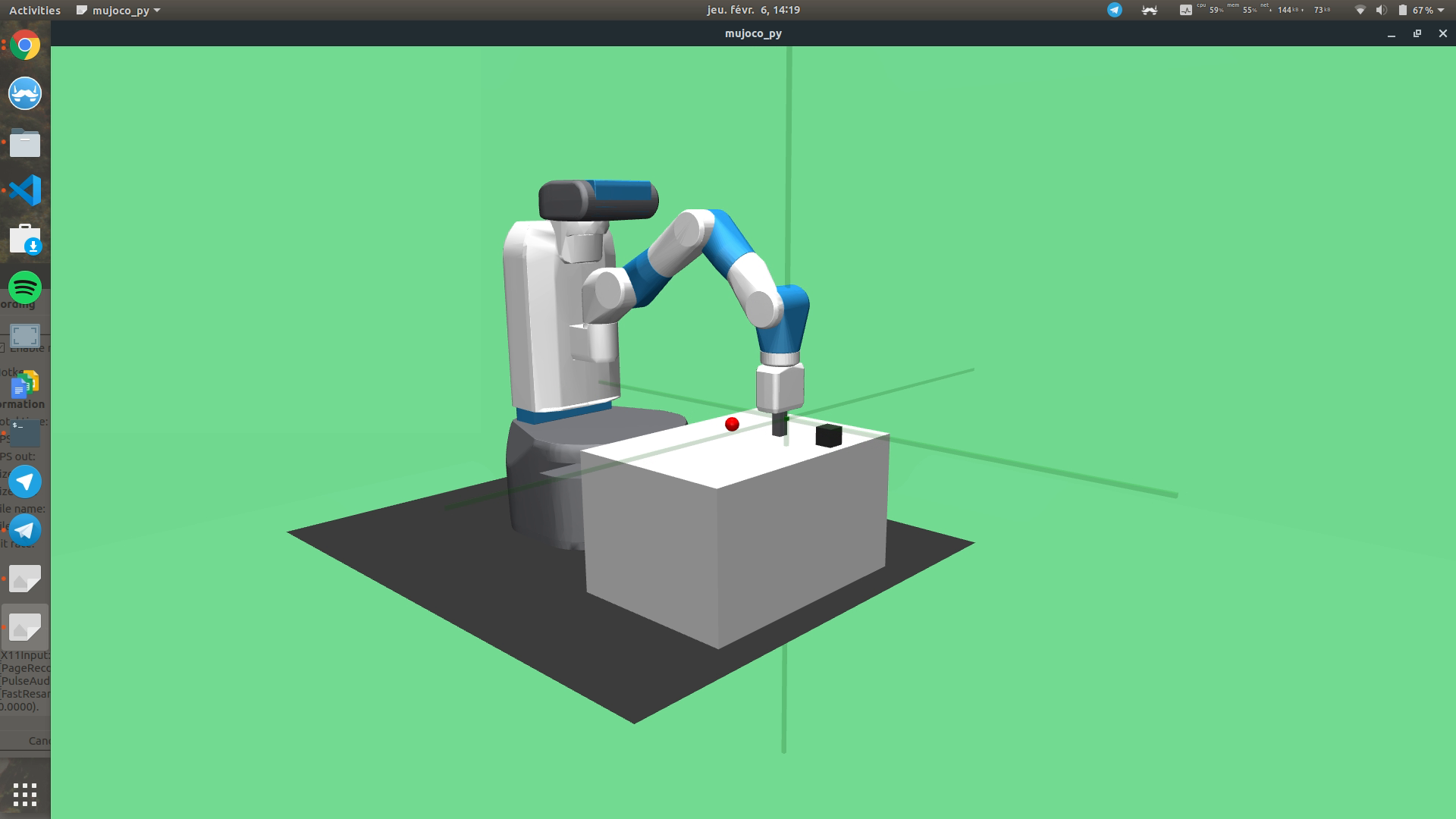}
\includegraphics[height=2.25cm,trim=320 50 550 120,clip]{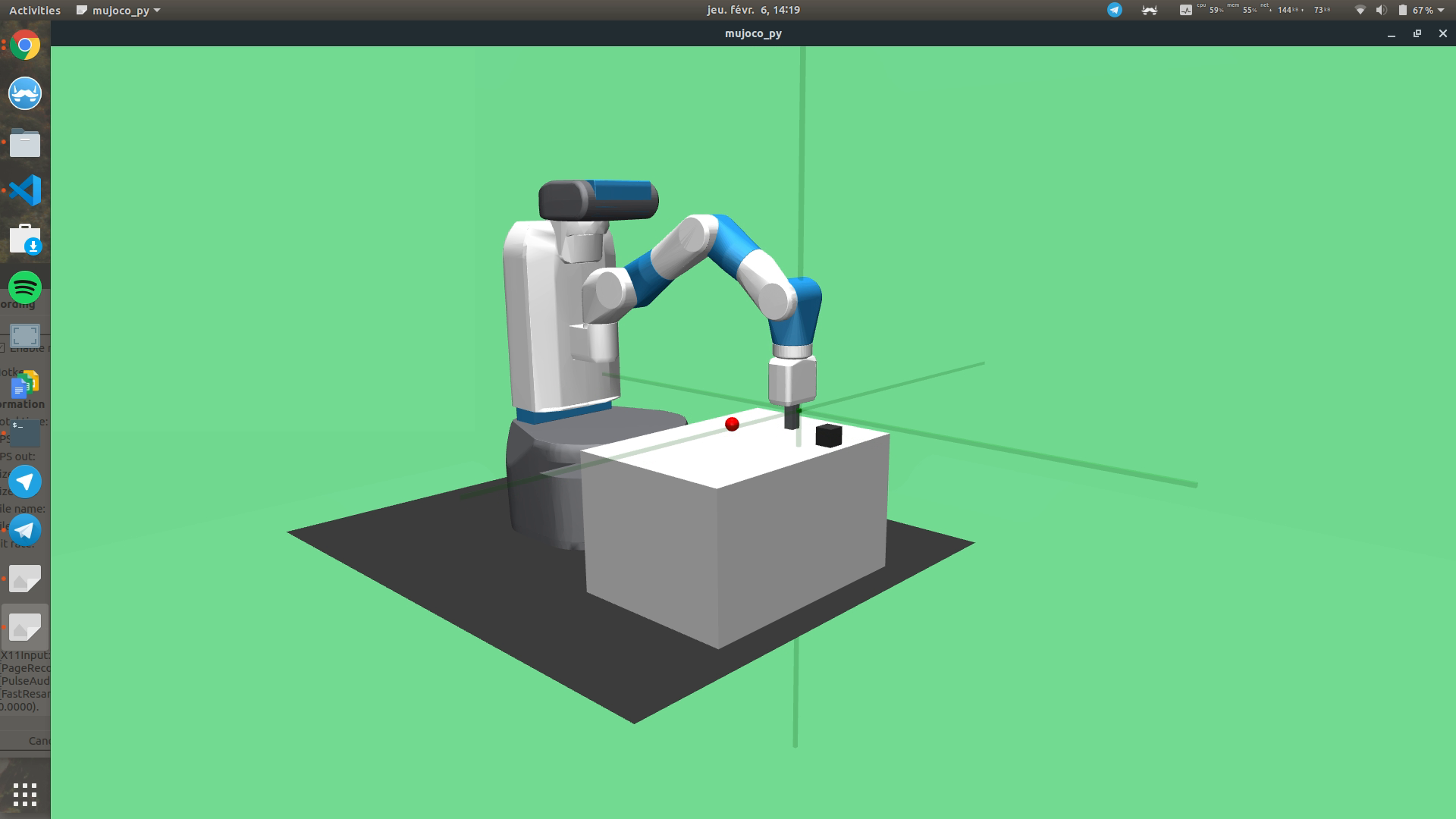}
}

\makebox[\textwidth][c]{%
\includegraphics[height=2.25cm,trim=320 50 550 120,clip]{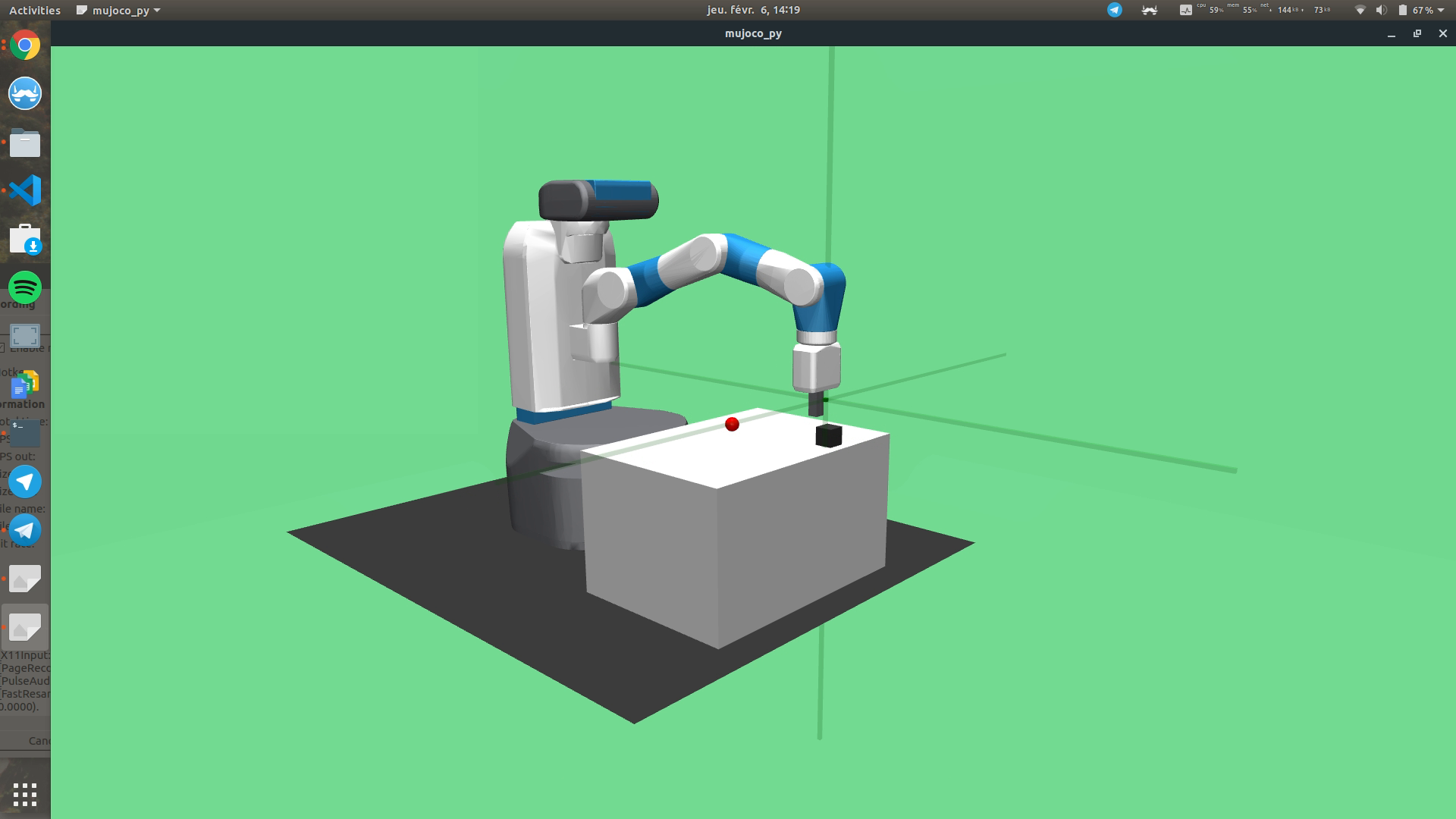}
\includegraphics[height=2.25cm,trim=320 50 550 120,clip]{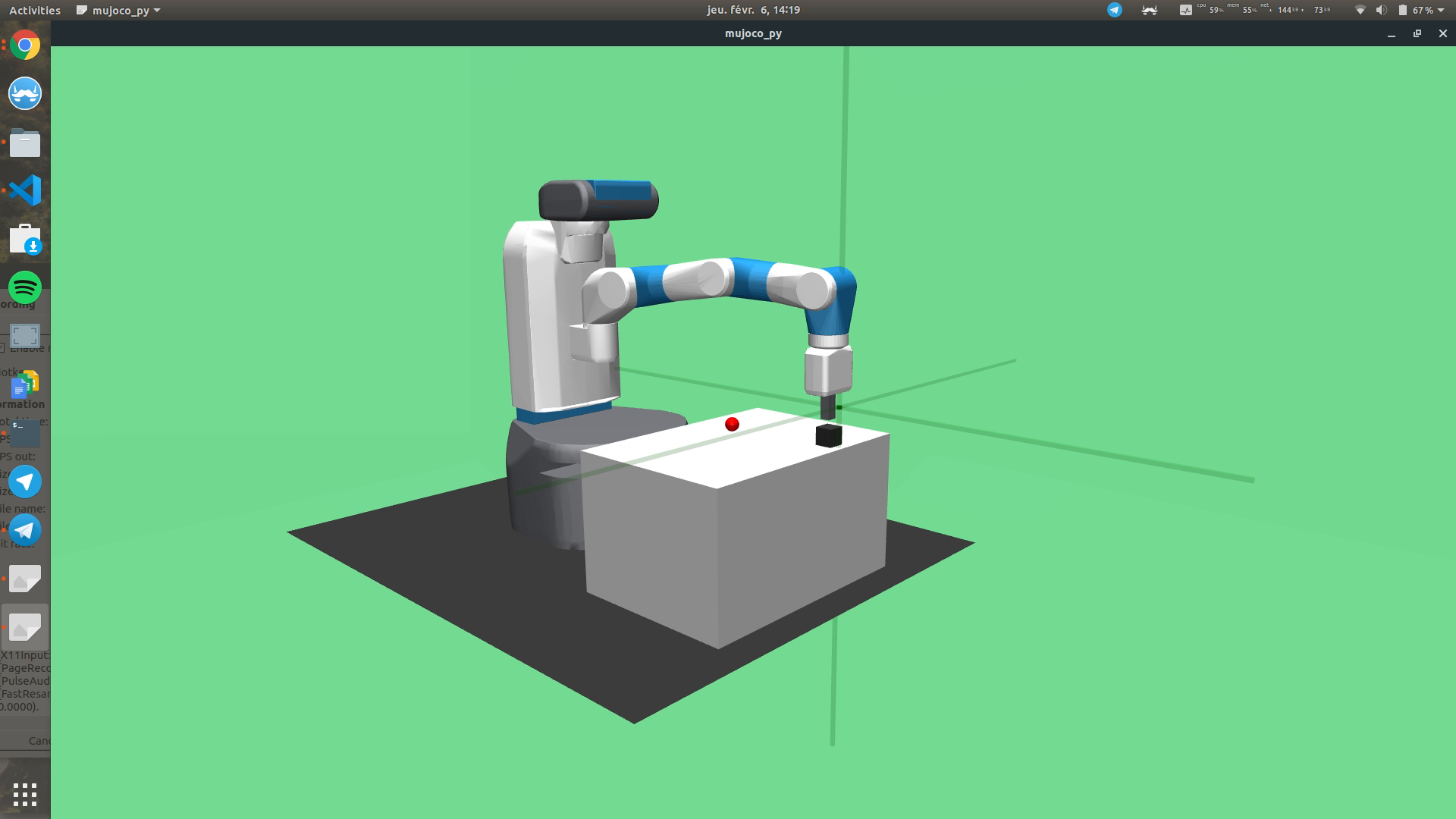}
\includegraphics[height=2.25cm,trim=320 50 550 120,clip]{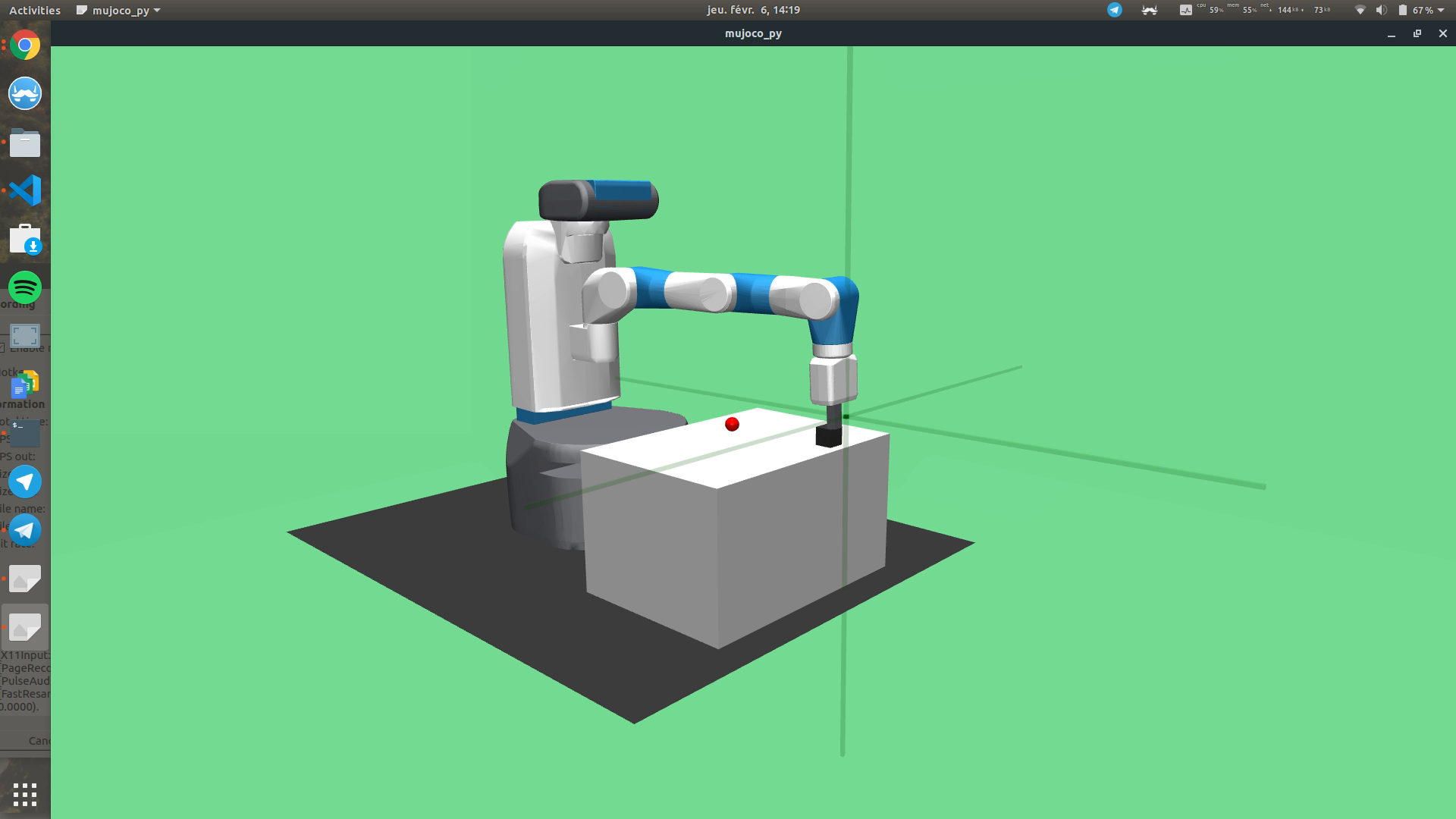}
}

\caption{Trajectory of MuJoCo's robotic arm when a new aimer is
  trained for the ``push'' task and the quasi-metric is kept from the
  ``pick and place'' task. The quasi-metric properly modeled how to
  move the arm at a desired location in contact with the black box,
  and needs minimal training to manipulate the box without gripper
  (see \S~\ref{sec:environment-tasks}).}\label{fig:mujoco-screenshots}
\end{minipage}

\end{figure}

\section{Method}

Let $\StateSpace$ be the state space and $\ActionSpace$ the action
space. We call {\bf goal} a subset $g \subset \StateSpace$ of the
state space, and a {\bf task} a set of goals $\GoalSpace \subset {\cal
  P}(\StateSpace)$. Many tasks can be defined in a given environment
with the same state and action spaces. Note that in the environments
we consider, the concrete definition of a task is a subset of the
state vector coordinates, and a goal is defined by the target values
for these coordinates.

Consider the robotic arm of the MuJoCo simulator \citep{mujoco2012},
that we use for experiments in \S~\ref{sec:experiments-mujoco}: The
state space $\StateSpace$ concatenates, among others, the position and velocity of the arm and the
location of the object to manipulate. Examples of tasks could be
``reach a certain position'', in which case a goal is a set of states
parameterized by a 3d position, where the position of the arm handle
is fixed but all other degrees of freedom are let free, ``reach a
certain speed'' where everything is let unconstrained but the handle's
speed, ``put the object at the left side of the table'', where
everything is free but one coordinate of the object location, and so
on.

For what follows, we also let
$(s_t, a_t, r_t) \in \StateSpace \times \ActionSpace \times \mathbb{R}, \ t = 1, \dots, T$
be a state / action / reward sequence.


\subsection{Planning Quasi-Metric}\label{sec:plann-quasi-metr}

Similarly to the distance between states proposed by
\citet{arxiv-1906.05253}, we explicitly introduce an
action-parameterized quasi-metric
\begin{equation}
f: \StateSpace^2 \times \ActionSpace \rightarrow \mathbb{R}_+
\end{equation}
such that $f(s, s'\!, a)$ is ``the minimum [expected] number of steps to
go from $s$ to $s'$ when starting with action $a$''.

We stress that it is a quasi-metric since it is not symmetric in most
of the actual planning setups. Consider for instance one-way streets
for an autonomous urban vehicle, irreversible physical
transformations, or inertia for a robotic task, which may make going
from $s_1=(x_1, v_1)$ to $s_2=(x_2, v_2)$ easy and the reciprocal
transition difficult.

Given an arbitrary target state $s'$, the update of $f$ should
minimize
\begin{equation}
\left( f(s_t, s_{t+1}, a_t) - 1 \right)^2 + \left( f(s_t, s'\!, a_t) - \left( 1 + \min_\alpha \tilde{f}(s_{t+1}, s'\!, \alpha) \right) \right)^2
\label{eq:pqm-update}
\end{equation}
where the first term makes the quasi-metric $1$ between successive
states, and the second makes it globally consistent with the best
policy, following Bellman's equation.
$\tilde{f}$ is a ``target model'', usually updated less frequently or
through a stabilizing moving average.

We implement the learning of the PQM with a standard actor/critic
structure \citep{lillicrap2015continuous}. First the PQM itself that plays the role of the critic
\begin{equation}
f_{w_f}: \StateSpace^2 \times \ActionSpace \rightarrow \mathbb{R}_+
\end{equation}
and an actor, which is either an explicit $\argmin$ in the case of a
finite set of actions, or a model
\begin{equation}
a_{w_a}: \StateSpace^2 \rightarrow \ActionSpace
\end{equation}
to approximate $a_{w_a}(s, s') \simeq \argmin_\alpha f_{w_f}(s, s',
\alpha)$ when dealing with a continuous action space $\ActionSpace.$

For training, given a tuple $(s_t, s_{t+1}, s'\!, a_t)$ we update
$w_f$ to reduce
\begin{multline}
\LL(w_f; s_t, s_{t+1}, s'\!, a_t, \tilde{w}_f) = \\ 
\left( f_{w_f}(s_t, s_{t+1}, a_t) - 1 \right)^2 + \left( f_{w_f}(s_t, s'\!, a_t)\!-\!\left(1\!+\!f_{\tilde{w}_f}(s_{t+1}, s'\!, a_{\tilde{w}_a}(s_{t+1} s')) \right) \right)^2\!\!\!
\label{eq:pqm-loss}
\end{multline}

and we update $w_a$ to reduce
\begin{equation}
f_{w_f}(s_t, s'\!, a_{w_a}(s_t, s')),
\end{equation}
so that $a_{w_a}$ gets closer to the choice of action at $s_t$ that
minimizes the remaining distance to $s'$.

\subsection{Aimer}\label{sec:aimer}

Note that while the quasi-metric allows to reach a certain state by
choosing at any moment the action that decreases the distance to it
the most, it does not allow to reach a more abstract ``goal'', defined
as a {\it set} of states. This objective is not trivial: the two
objects are defined at completely different scales, the latter
possibly ignoring virtually all the degrees of freedom of the former.

Hence, to use the PQM to actually reach goals, a key element is
missing to pick the ``ideal state'' that (1) is in the goal but also
(2) is the easiest to reach from the state currently occupied.
For this purpose we introduce the idea of aimer (see
figure~\ref{fig:aimer}) which, given a set of goals $\GoalSpace
\subset {\cal P}(\StateSpace)$, is of the form
\begin{equation}
h^{\GoalSpace}: \StateSpace \times \GoalSpace \rightarrow \StateSpace
\end{equation}
and is such that $h^{\GoalSpace}(s, g)$ is the ``best'' target state, that is the
state in $g \in \GoalSpace$ closest to $s$:
\begin{equation}
\forall s, g \in \StateSpace \times \GoalSpace, \ h^{\GoalSpace}(s, g) = \argmin_{s' \in g} \min_a f(s, s'\!, a).
\label{eq:aimer-optim}
\end{equation}

The key notion in this formulation is that we can have multiple
aimers dedicated to as many goal spaces, that utilize the same
quasi-metric, which is in charge of the heavy lifting of
``understanding'' the underlying dynamics of the environment.

We follow the idea of the actor for the action choice, and do not
implement the aimer by explicitly solving the system of equation
\ref{eq:aimer-optim} but introduce a parameterized model
\begin{equation}
h_{w_h}: \StateSpace \times \GoalSpace \rightarrow \StateSpace.
\end{equation}
For training, given a pair $(s, g) \in \StateSpace \times \GoalSpace$
we update $w_h$ to reduce 
\begin{equation}
f_{w_f}(s, h_{w_h}(s, g), a_{w_a}(s, h_{w_h}(s, g))) \\
+ \lambda_1 \, d(h_{w_h}(s, g), g)
+ \lambda_2 \, v(h_{w_h}(s, g)).
\end{equation}

The first term is an estimate of the objective of the problem
(\ref{eq:aimer-optim}), that is the distance between $s$ and
$h_{w_h}(s, g)$, where the actor's prediction plays the role of the
$\min$ of the original problem.

The second term is a penalty replacing the hard constraints of
(\ref{eq:aimer-optim}) with a distance $d$ to a set. That latter
distance is in practice a $L_2$ norm over a subset of the state's
coordinates. We come back to this with more details in
\S~\ref{sec:experiments}.

The third term $v$ is a penalty for imposing the validity of the
state, for instance ensuring that speed or angles remain in valid
ranges.

The resulting policy combines the actor $a$ and the aimer $h$. Given
the current state $s$ and the goal $g$, the chosen action is $a(s,
h(s, g))$.

\section{Experiments}\label{sec:experiments}

We have validated our approach experimentally in
PyTorch \citep{PyTorch2019NIPS}, on two standard environments: the
bit-flipping problem (see \S~\ref{sec:bitflip-exp}), known to be
particularly challenging to traditional RL approaches relying on
sparse rewards, and the MuJoCo simulator (see
\S~\ref{sec:experiments-mujoco}), which exhibits some key difficulties
of real robotic tasks. Our software to reproduce the experiments will
be available under an open-source license at the time of publication.

To ensure that the benefits of PQM are complementary to
state-of-the-art strategies making a smart re-use of episodes, we have compared our approach to the performance of DQN+HER from \cite{arxiv-1707.01495} that
we name here solely DQN for the bit-flipping problem and
DDPG+HER from \cite{plappert2018multi} that we name here solely DDPG for the Mujoco simulator. Hence all the algorithms we consider in the experimental part, both baselines and our PQM, are using HER.

The experimental results presented in this section were obtained with roughly 250 vCPU cores for one month, which is far less than the requirement for some state-of-the-art results. It forced us to only coarsely adapt configurations optimized in previous works for more classical and consequently quite different approaches.

\subsection{Bit-flip}\label{sec:bitflip-exp}

\subsubsection{Environment and tasks}\label{sec:bitflip-environment-tasks}

The state space for this first environment is a Boolean vector of $n$
bits, and there are $n$ actions, each switching one particular bit of
the state. We fix $n=30$ and define two tasks, corresponding to
reaching a target configuration respectively for the first $15$ bits
and the last $15$ bits. A goal in these tasks is defined by the target
configuration of $15$ bits.

The difficulty in this environment is the cardinality of the state
space, the lack of geometrical structure, and the average time it
takes to go from any state to any other state under a random policy.

To demonstrate the transferability of the quasi-metric, we also
consider a task defined by the 15 first bits, with transfer from a
task defined by the 15 last bits. After training an aimer and
quasi-metric on one, we train a new aimer on the other but keep and
fine-tune the parameters $w_f$ of the quasi-metric.

\subsubsection{Network architectures and training}

The critic is implemented as $f(s, s', a) = \phi_a(s, s')$, where
$\phi$ is a ReLU MLP with $2n = 60$ input units, one hidden layer with
$256$ neurons, and $n =30$ outputs.  As indicated in
\S~\ref{sec:plann-quasi-metr}, the actor for this environment is an
explicit $\argmin$ over the actions. The aimer $h$ is implemented
also as a ReLU MLP with $n + n/2 = 45$ input units corresponding to
the concatenation of a state and a goal definition, one hidden layer
with $256$ neurons, and $n = 30$ output neurons, with a final sigmoid
non-linearity.

The length of an episode is equal to the number of bits in the goal,
which is twice the median of the optimal number of actions.
We kept the meta-parameters as selected by \citet[appendix
  B]{plappert2018multi}, and chose $\lambda_1 = 100$. There is no term
imposing the validity of the aimer output, hence no parameter
$\lambda_2$.

Following algorithm S.1, we train for $500$ epochs,
each consisting of running the policy for $16$ episodes and then
performing $40$ optimization steps on minibatches of size $256$
sampled uniformly from a replay buffer consisting of $10^6$
transitions. We use the ``future'' strategy of HER for the selection
of goals \cite{arxiv-1707.01495}. We update the target networks after
every optimization step using the decay coefficient of $0.95$.

\begin{figure*}[ht!]
  \hspace*{-10pt}
  \includegraphics[scale=0.455]{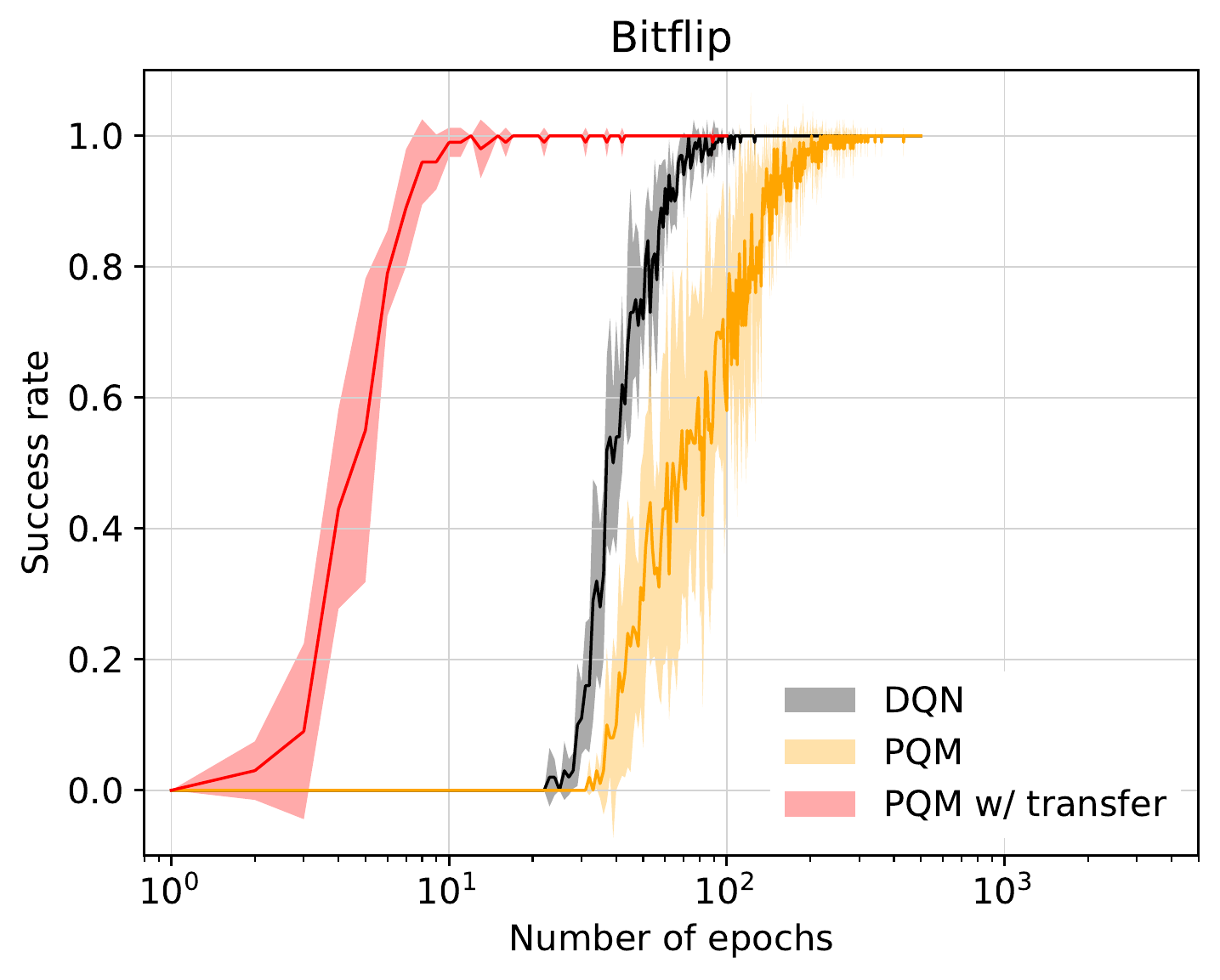}
  \includegraphics[scale=0.455]{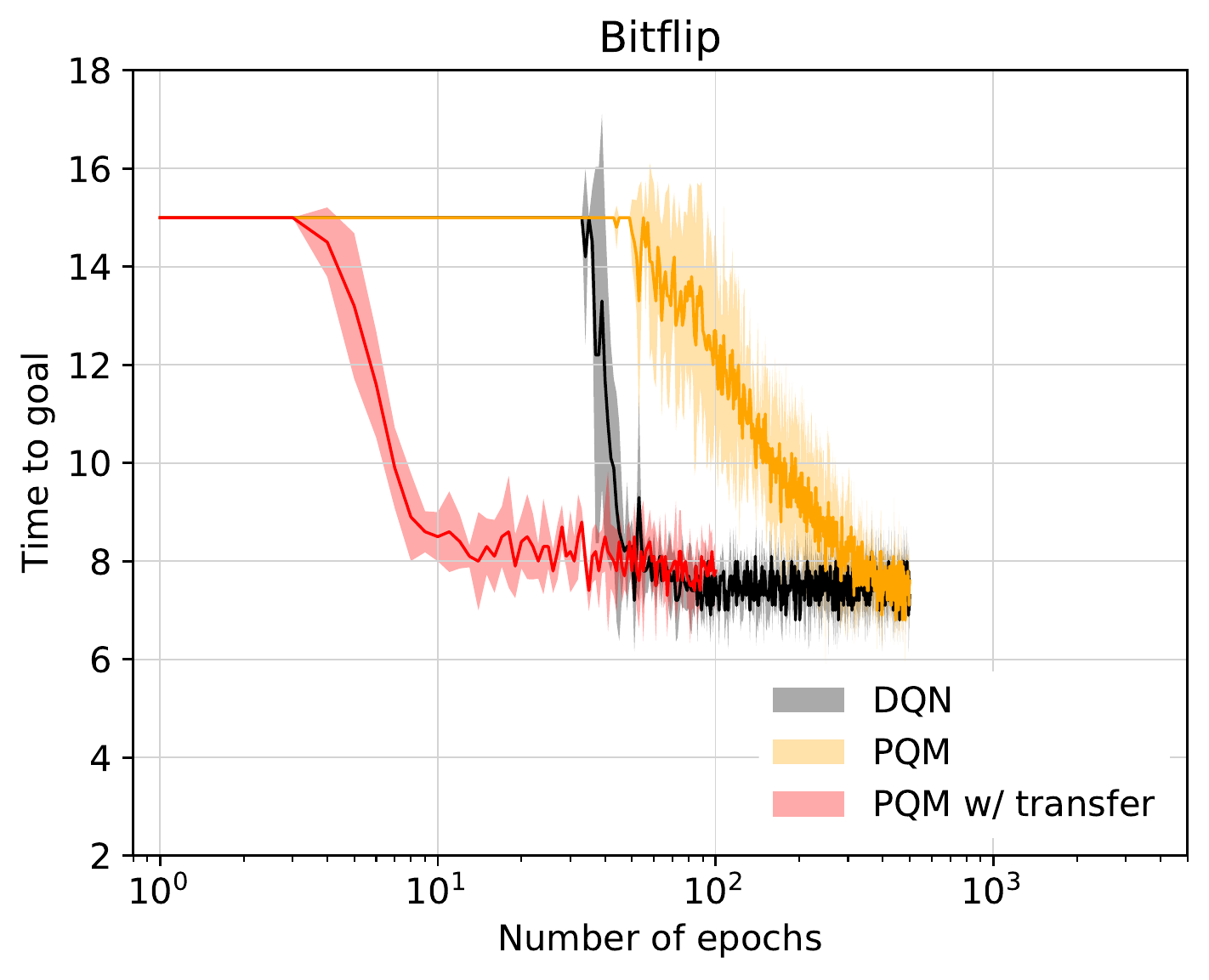}
  \caption{Empirical mean success rate (left) and median time to goal
    (right) with $\pm$ one standard deviation confidence interval on
    the bit-flip task (see \S~\ref{sec:bitflip-environment-tasks}) of
    three different algorithms: a standard deep Q-learning (DQN), our
    approach that combines a planning quasi-metric with an aimer
    (PQM), and the same with training of an aimer from scratch and
    transfer of the quasi-metric trained on a task where the bits to
    match to reach the goals were different (PQM w/ transfer). That
    latter curve shows a boost in early training thanks to the
    pre-trained quasi-metric.}
  \label{fig:bitflip-perf}
\end{figure*}

\begin{figure*}[ht!]
  \hspace*{\stretch{1}}
  \includegraphics[scale=0.455]{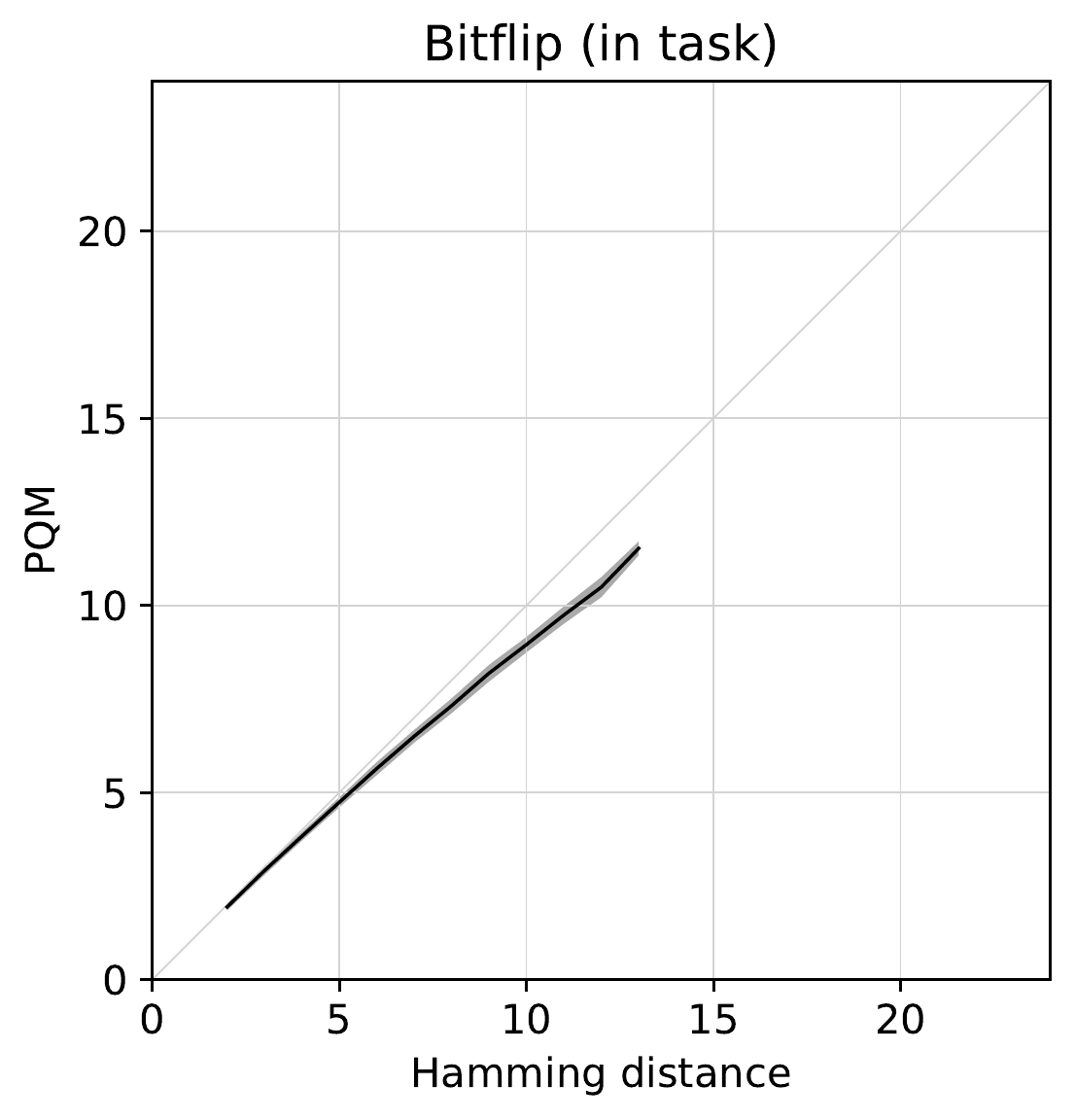}
  \hspace*{\stretch{1}}
  \includegraphics[scale=0.455]{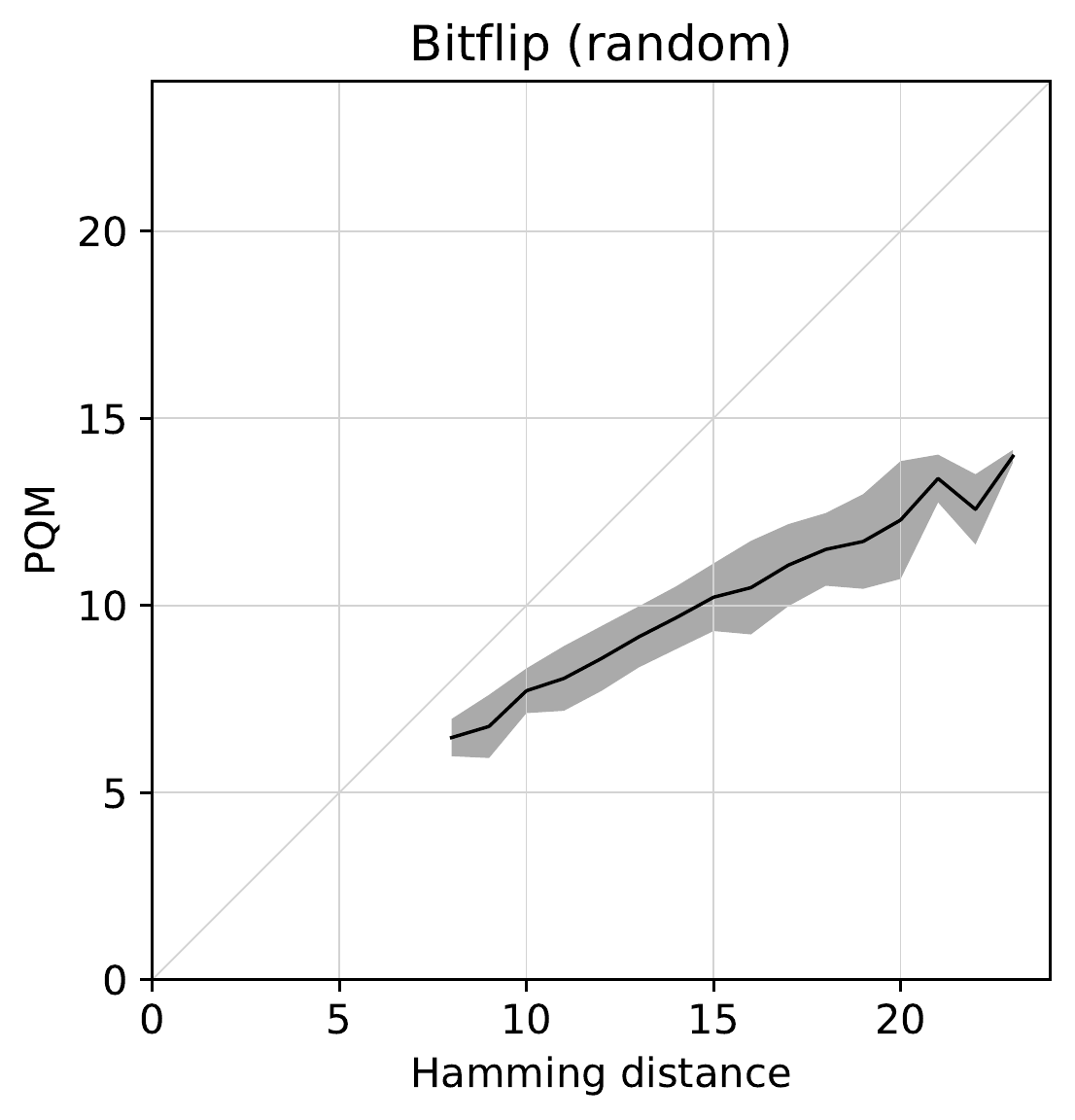}
  \hspace*{\stretch{1}}
  \caption{Accuracy of the quasi-metric estimate on the bit-flip task (see
    \S~\ref{sec:bitflip-environment-tasks}).
We plot here the empirical mean and $\pm$ one standard deviation confidence
interval of the PQM estimate $\min_a f(s, s', a)$ vs. the true
distance, which here is the number of differing bits, hence the
Hamming distance. The left figure is obtained when the starting state
and the goal state differ only on the bits that are relevant to the
task, hence are consistent with the aimer $h$, and the right figure
when both are taken at random.}\label{fig:bitflip-metric}
\end{figure*}

\subsubsection{Results}

The experiments in the bit-flip environment show the advantage of
using a planning quasi-metric. As shown on figure
\ref{fig:bitflip-perf}, the training is successful and the combination
of the quasi-metric and the aimer results in a policy similar to that
of the standard DQN, both in terms of success rate and in terms of
time to goal. It also appears that, while this model is slightly
harder to train on a single task compared to DQN, it provides a great
performance boost when transferring the PQM between tasks: since the
quasi-metric is pre-trained, the training process only needs to train
an aimer, which is a simple model, and fine-tune the quasi-metric.

It is noteworthy that due to limited computational means, we kept
essentially the meta-parameters of the DQN setup of
\cite{arxiv-1707.01495}, and as such the comparison is biased to DQN's advantage.

Figure \ref{fig:bitflip-metric} gives a clearer view of the accuracy
of the metric alone. We have computed after training the value of
$\min_a f(s, s', a)$ for pairs of starting states / target states
taken at random, and compared it to the ``true'' distance, which
happens to be in that environment the Hamming distance, that is the
number of bits that differ between the two.

For the tasks in this environment, the aimer $h$ predicts a target
state whose bits that matter for the task are the goal configuration,
and the others are unchanged from the starting state, since this
corresponds to the shortest path. Hence we considered two groups of
state pairs: Either ``in task'', which means that the two states are
consistent with the aimer prediction in the task, and differ only on
the bits that matter for the task, or ``random'' in which case they
are arbitrary, and hence may be inconsistent with
the biased statistic observed during training.

The results show that the estimate of the quasi-metric is very
accurate on the first group, less so on the second, but still strongly
monotonic. This is consistent with the transfer providing a
substantial boost to the training on a new task.

\subsection{MuJoCo}\label{sec:experiments-mujoco}
\subsubsection{Environment and tasks}\label{sec:environment-tasks}

For our second set of experiments, we use the ``Fetch'' environments
of OpenAI gym \citep{gymOpenAI1606.01540} which use the MuJoCo physics
engine \citep{mujoco2012}, pictured in figure
\ref{fig:mujoco-screenshots}, and are described in details by
\citet[section 1.1]{plappert2018multi}. If left unspecified the
details of our experiments are the same as indicated by \citet[section
  1.4]{plappert2018multi}, and \citet[appendix A]{arxiv-1707.01495}.

We consider two tasks: ``push'', where a box is placed at random on
the table and the robot's objective is to move it to a desired
location also on the table, without using the gripper, and ``pick and
place'', in which the robot can control its gripper, and the desired
location for the box may be located above the table surface.

To demonstrate the transferability of the quasi-metric in this
environment, we also consider the ``push'' task with transfer from
``pick and place'': after training an aimer and the quasi-metric on
the latter, we train a new aimer on the former, but keep the
parameters $w_f$ and $w_a$ as starting point for the quasi-metric.

Note that our implementation of DDPG outperforms the results obtained by \citet{plappert2018multi}. This is due to the addition of an extra factor in the loss of the critic which sets the target Q value to 0 whenever the goal is reached. In our approach, the metric does not entail at all the notion of a goal, hence this factor was omitted.

\begin{figure*}[ht!]
  \hspace*{-10pt}
  \includegraphics[scale=0.455]{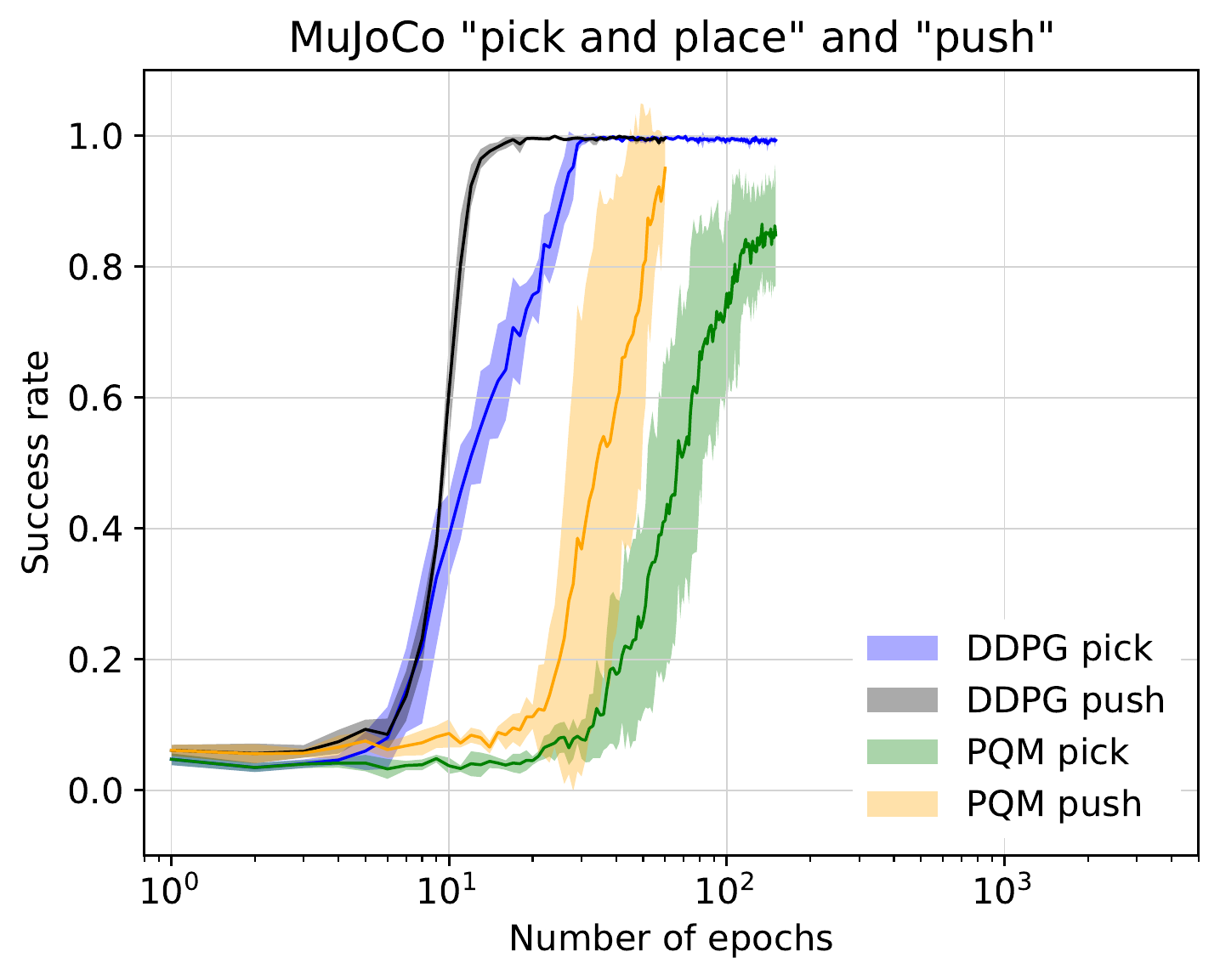}
  \includegraphics[scale=0.455]{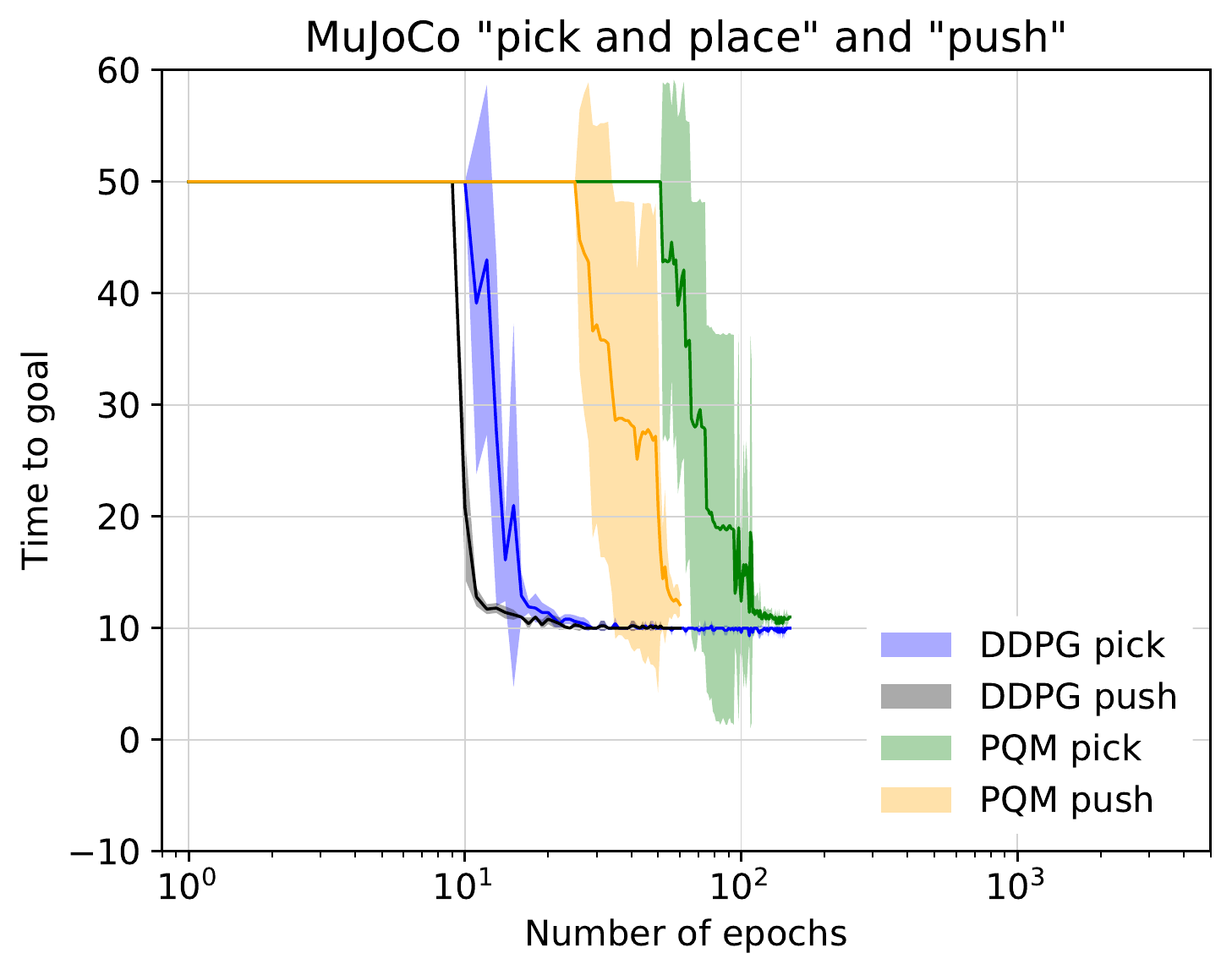}
  \caption{Empirical mean success rate (left) and median time to goal (right) with $\pm$ one standard deviation confidence interval on the MuJoCo
  ``pick and place'' and ``push'' tasks (see
  \S~\ref{sec:environment-tasks}). We compare the performance of the
  Deep Deterministic Policy Gradient (DDPG), with our approach that
  combines a planning quasi-metric with an aimer (PQM). These curves
  show that although the metric is harder to learn than the policy
  alone, the joint learning of the two models is
  successful.} \label{fig:mujoco-perf}
\end{figure*}

\begin{figure*}[ht!]
  \hspace*{-10pt}
  \includegraphics[scale=0.455]{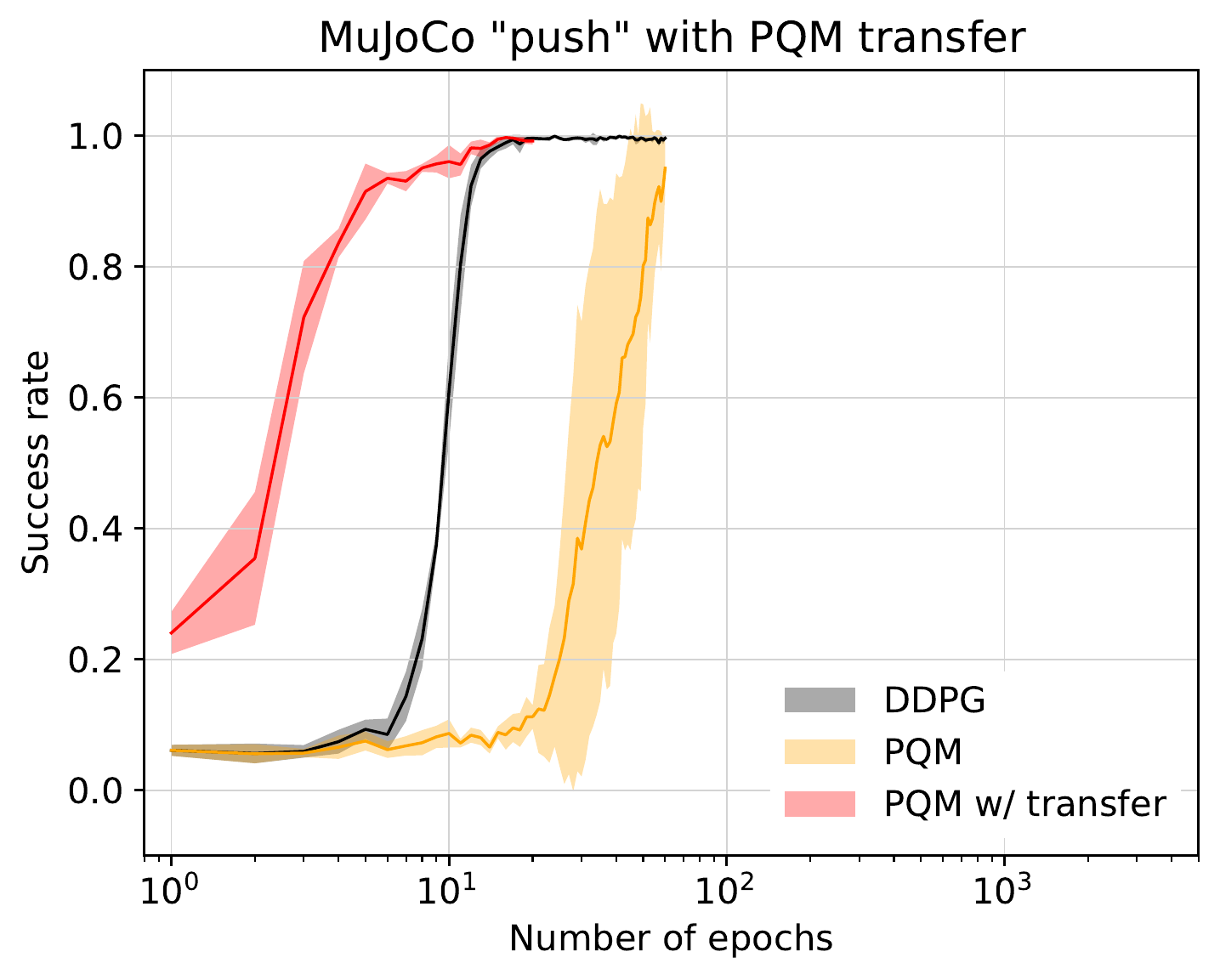}
  \includegraphics[scale=0.455]{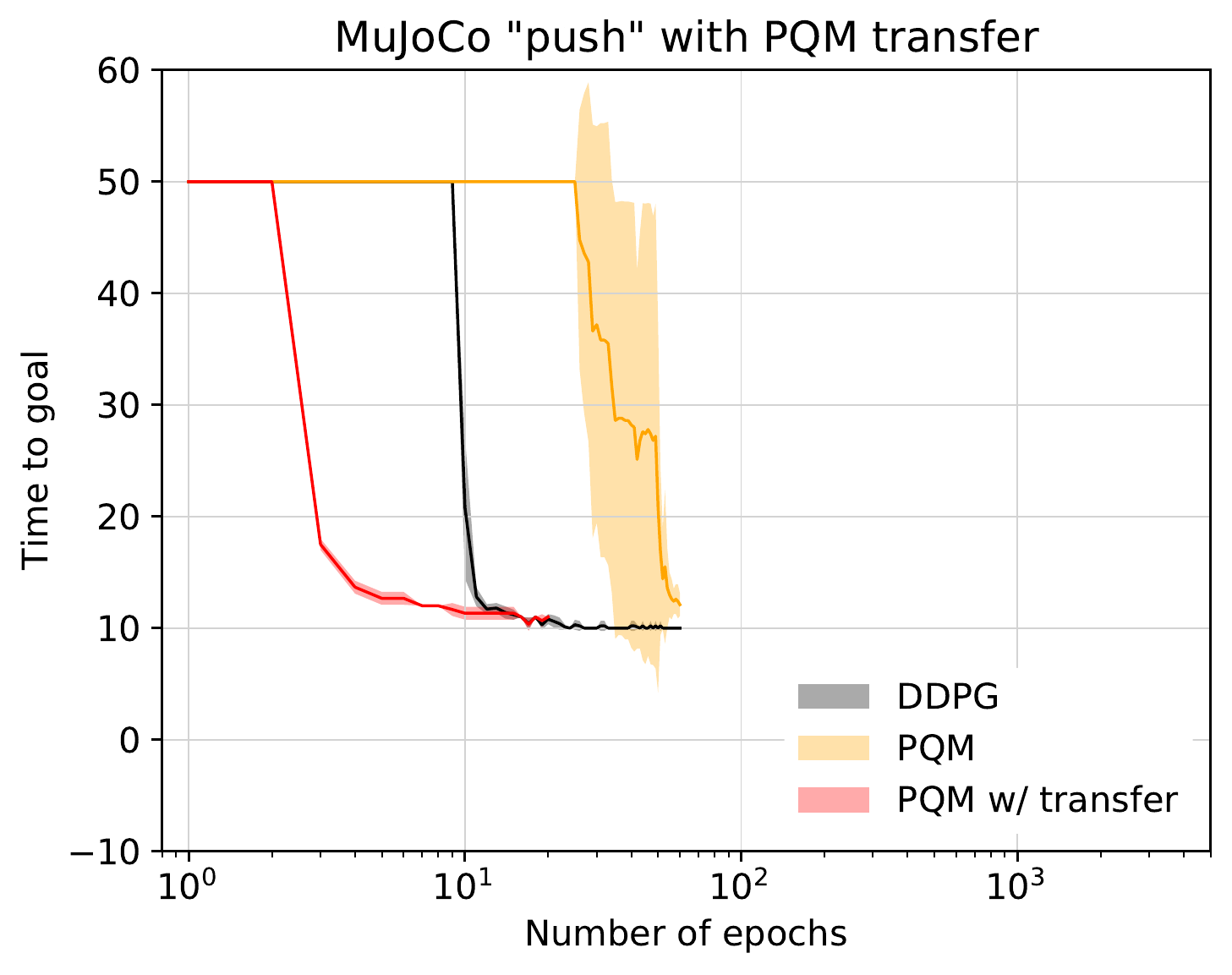}
  \caption{Empirical mean success rate (left) and median time to goal
    (right) with $\pm$ one standard deviation confidence interval on
    the MuJoCo ``push'' task (see \S~\ref{sec:environment-tasks}).  We
    compare the performance of the Deep Deterministic Policy Gradient
    (DDPG), with our approach that combines a planning quasi-metric
    with a aimer (PQM), and the same with training of an aimer from
    scratch and transfer of the quasi-metric trained on the ``pick and
    place'' task (PQM w/ transfer). That latter curve shows a boost in
    early training thanks to the pre-trained
    quasi-metric.} \label{fig:mujoco-transfer-perf}
\end{figure*}

\begin{figure*}[ht!]
  \hspace*{\stretch{1}}
  \includegraphics[scale=0.455]{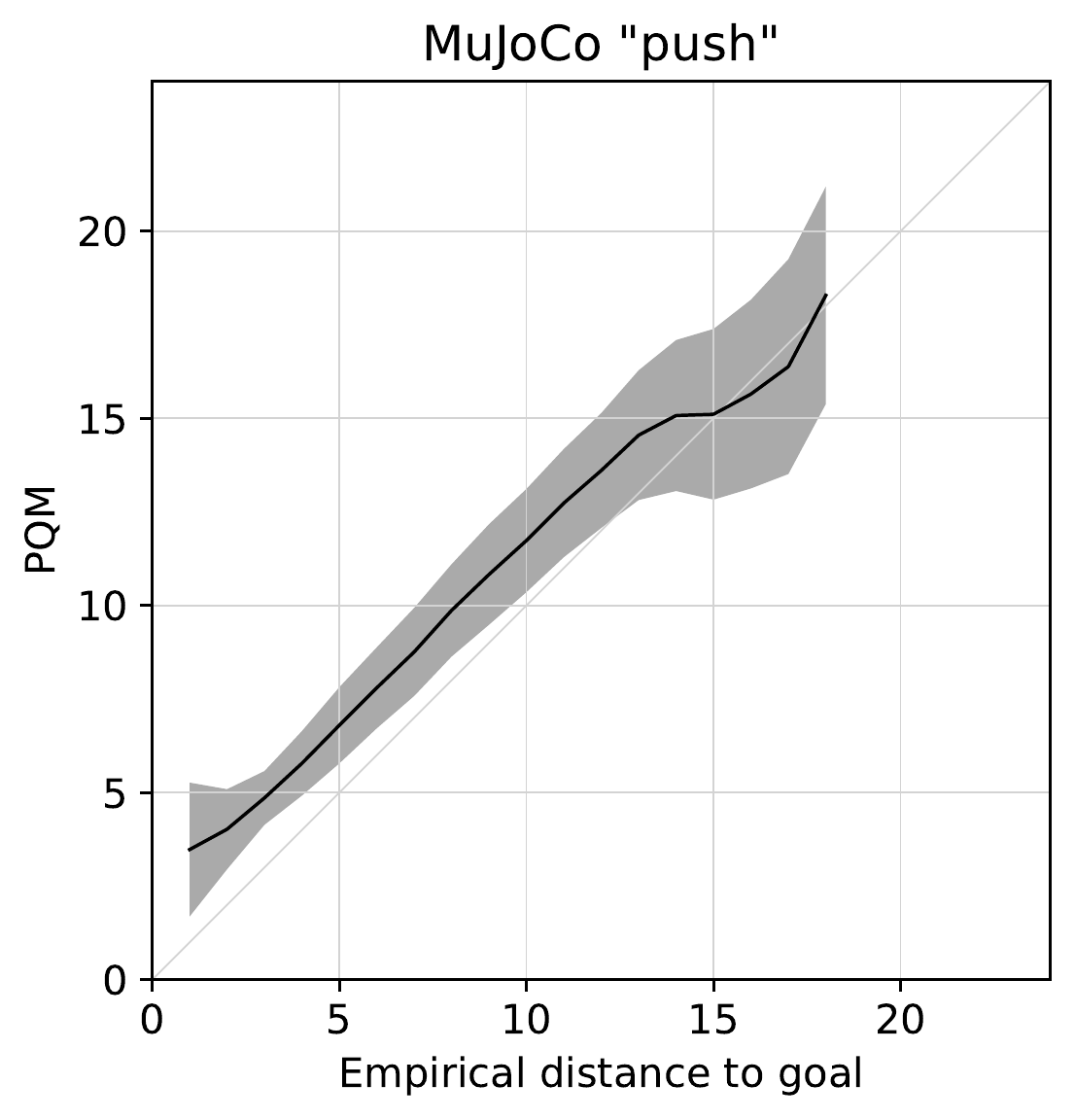}
  \hspace*{\stretch{1}}
  \includegraphics[scale=0.455]{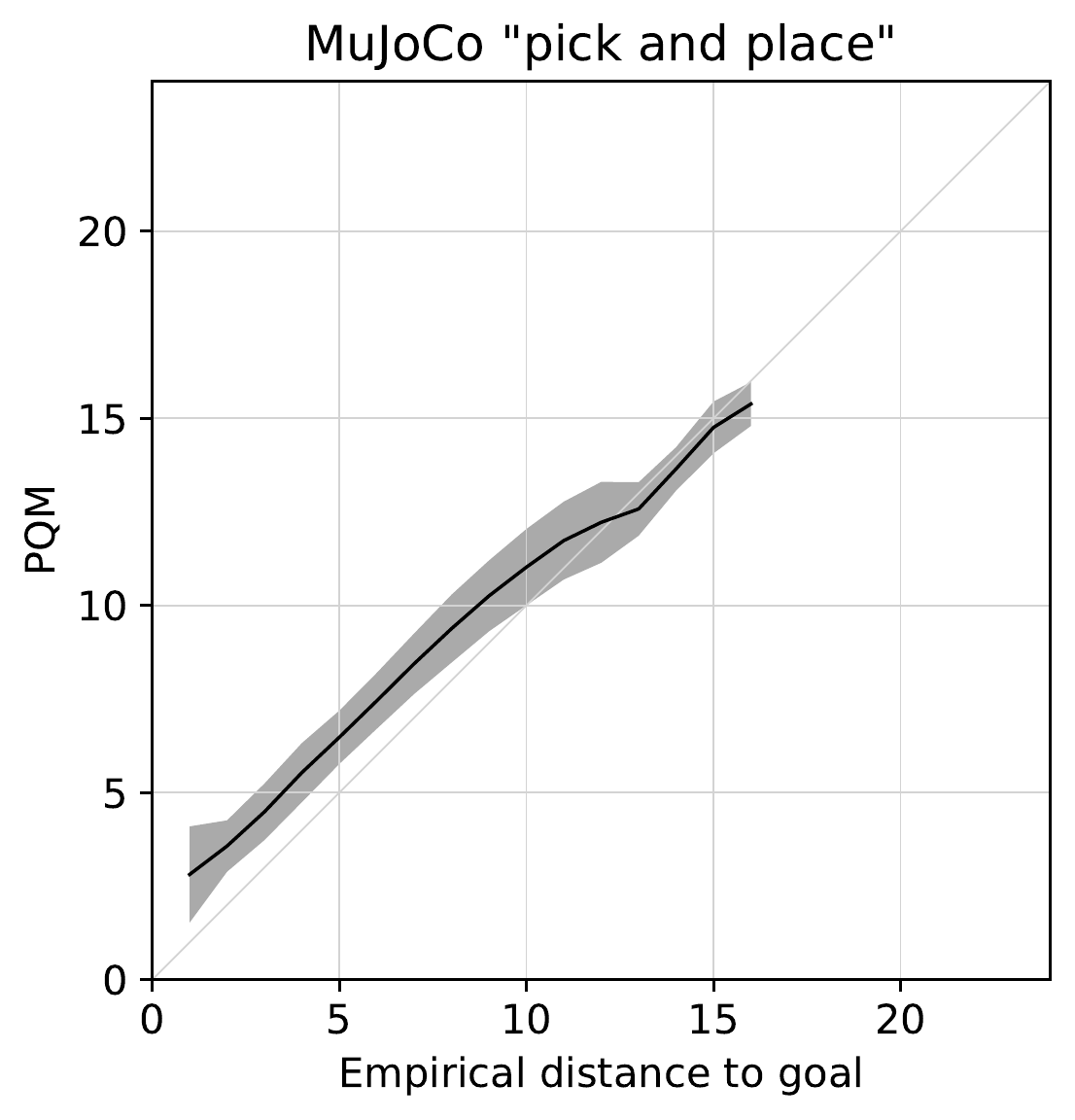}
  \hspace*{\stretch{1}}
  \caption{Accuracy of the quasi-metric on the MuJoCo ``push'' and
  ``pick and place'' tasks (see \S~\ref{sec:environment-tasks}). We
  plot here the empirical mean and $\pm$ one standard deviation confidence
  interval of the PQM estimate $f(s, s', a(s, s'))$ at the beginning
  of a successful episode, where $s'$ is the target goal as estimated
  by the aimer vs. the actual number of steps it took to reach
  $s'$.}\label{fig:mujoco-metric}
\end{figure*}

\subsubsection{Network architectures and training}

In what follows let $d_s=25$ be the dimension of the state space
$\StateSpace$, $d_a = 4$ the dimension of the action space
$\ActionSpace$, $d_g = 3$ the dimension of the goal parameter, which
corresponds to the desired spatial location of the manipulated box.

The critic $f$ is implemented with a ReLU MLP, with $2 d_s + d_a$
input units, corresponding to the concatenation of two states and an
action, three hidden layers with $512$ units each, and a single output
unit. The actor $a$ is a ReLU MLP with $2 d_s$ input units, three
hidden layers with $512$ units each, and $d_a$ output units with
$\tanh$ non-linearity. As \cite{plappert2018multi}, we also add a
penalty to the actor's loss equal to the square of the output layer
pre-activations. Finally the aimer $h$ is a ReLU MLP with $d_s +
d_g$ input units, three hidden layers of $256$ units, and $d_s$ output
units.

Following the algorithm S.1, for ``pick and place''
and ``push with transfer'', we train for $150$ epochs. For ``push'',
we train for $60$ epochs. As in \cite{plappert2018multi}, each epoch
consists of $50$ cycles, and each cycle consists of running the policy
for $2$ episodes and then performing $40$ optimization steps on
minibatches of size $256$ sampled uniformly from a replay buffer
consisting of $10^6$ transitions. We use the ``future'' strategy of
HER for the selection of goals \citep{arxiv-1707.01495}, and update the
target networks after every cycle using the decay coefficient of
$0.95$.

We kept the meta-parameters as selected in \citet[appendix
  B]{plappert2018multi}, with an additional grid search over the
number of hidden neurons in the actor and critic models in $\{ 256,
512 \}$, $\lambda_1$ in $\{ 50, 100, 250, 500, 750, 1000 \}$, and
$\lambda_2$ in $\{ 0, 1, 5, 10, 50, 100, 250, 500, 750, 1000 \}$.

For each combination, we trained a policy on the ``push'' task, and
eventually selected the combination with the highest rolling median
success rate over $10$ epochs, resulting in $512$ hidden neurons,
$\lambda_1 = 500$, and $\lambda_2 = 50$. Note that additional trials tuning the learning rate of the models did not yield significant improvements.

All hyperparameters are described in greater detail by \citet{arxiv-1707.01495}.


\subsubsection{Results}

The results obtained in this environment confirm the observations from
the bit-flip environment. Figure \ref{fig:mujoco-perf} shows that the
joint training of the PQM and aimers works properly and results in a
policy similar to that obtained with the DDPG approach. Once again, the PQM is slightly harder to train on a single task, in part due to the limited meta-optimization we could afford that favors the baseline, and in part due to the difficulty of learning the metric, which is a more complicated functional. However, our approach outperforms DDPG when the aimer alone has to be trained from scratch, and the PQM  is transferred between tasks.

Figures \ref{fig:mujoco-screenshots} and
\ref{fig:mujoco-transfer-perf} show the advantage of using the PQM to
transfer knowledge from a task to another. Even though the two tasks
are quite different, one using the gripper and moving the object in
space while holding it, and the other moving only by contact in the
plane, the quasi-metric provides an initial boost in the training by
providing the ability to position the arm.

Finally, Figure \ref{fig:mujoco-metric} shows that the estimate of the
quasi-metric accurately reflects the actual distance to the goal
state.


\section{Related works}

The idea of a goal-conditioned policy combined with a
constant negative reward, which results in an
accumulated reward structure having the form of the [opposite of] the
distance to the goal~\citep{Kaelbling93learningto} has seen a strong renewal of interest recently. \citet{arxiv-1906.05253} explicitly consider a
distance between states, \cite{DBLP:journals/corr/abs-1907-08225} utilize a learned distance function to efficiently
optimize a goal-reaching policy and \cite{DBLP:journals/corr/abs-1809-09318} optimize a value function between states, which accounts for the distance function, based on the triangular inequality. As such, these works make use of a metric between states but do not use the notion of aimer in order to generate a state-goal to reach from the current state. While \cite{nasiriany2019planning} and \cite{DBLP:journals/corr/abs-1901-00943} use a similar concept to the aimer, it can in practice only generate nearby goals from the current state and there is no explicit transferability through a shared distance metric among different tasks.

The idea of transferring models has been applied to reinforcement
learning~\cite{taylor2009}, with recent successes using a single model
that mimics specialized expert actors on individual
tasks~\cite{parisotto2016}. The key issue of bringing several models
to a common representation is tackled by normalizing contributions of
the different tasks~\cite{arxiv-1809.04474}. This is in contrast with
our proposal, which explicitly leverages being in a similar
environment, and corresponds in our view to a more realistic robotic
setup for which the embodiment is fixed.

More recently, \citet{learningtoplanchen} address the transferability problem through the use of attention and planning modules. An embedding is learnt to go from a high dimensional continuous state space into a low dimensional discrete one in order to facilitate the planning process. Prior training of the policy and value function approximators is leveraged to reduce the number of required samples for solving new tasks. In a similar setting, \citet{rl-rtt} propose to learn a transferable obstacle avoiding state-to-state policy with evolutionary algorithms. From the rollouts generated, a time to reach estimator is learnt and used to grow a tree of nearby states to attain during the actual planning.

\section{Conclusion}

We have proposed to address the action selection problem by modeling
separately the estimation of a
target state given a goal and a quasi-metric between states. Experiments show that these two models can be trained jointly to get an efficient policy and that this approach supersedes robust baselines. As also illustrated in the experiments, the core
advantage is that this decomposition moves the bulk of the modeling to
the quasi-metric, which can be trained across tasks, with a dense
feedback from the environment, while the aimer can be trained very quickly from a small number of episodes.

By disentangling two very different aspects of the planning, this
decomposition is very promising for future extensions. The aimers are easier to learn and may be improved with
a specific class of regressors taking advantage
of a coarse-to-fine structure: your final destination can be initially
coarsely defined and refined along your way. The
quasi-metric handles the difficulty of learning a global structure
known only through local interactions but is potentially amenable to
the triangular inequality, clustering methods, and dimension
reduction.


\section*{Broader Impact}

For the past decade, Deep Reinforcement Learning has been paving the way towards Artificial General Intelligence. It is our belief that transfer learning and shared representations constitute a step in that direction.

Nonetheless, Deep Reinforcement Learning requires tremendous computational resources that only a few companies and institutions can afford. Our approach allows one to speedup the learning process thus lowering the entry barrier to practical applications and new research. Besides, power consumption has a direct environmental cost.

\section*{Acknowledgments}

Experiments were supported in part by Amazon through
the AWS Cloud Credits for Research Program.

\bibliography{neurips_workshop_pqm}
\bibliographystyle{abbrvnat}

\end{document}